\documentclass[journal]{IEEEtran}
\usepackage{graphicx,times,amsmath,amssymb,subfigure,stfloats,threeparttable,booktabs,bm,multirow,cite}
\usepackage[lined,ruled,commentsnumbered]{algorithm2e}
\allowdisplaybreaks

\makeatletter

\newcommand{\Rmnum}[1]{\expandafter\@slowromancap\romannumeral #1@}
\newcommand{\tabincell}[2]{\begin{tabular}{@{}#1@{}}#2\end{tabular}}
\makeatother

\begin{document}

\title{Optimize TSK Fuzzy Systems for Regression Problems: Mini-Batch Gradient Descent with Regularization, DropRule, and AdaBound (MBGD-RDA)}

\author{Dongrui~Wu, Ye Yuan, Jian~Huang, and Yihua Tan
\thanks{D.~Wu, Y.~Yuan, J.~Huang and Y.~Tan are with the Ministry of Education Key Laboratory of Image Processing and Intelligent Control, School of Artificial Intelligence and Automation, Huazhong University of Science and Technology, Wuhan, China. Email: drwu@hust.edu.cn, yye@hust.edu.cn, huang\_jan@hust.edu.cn, yhtan@hust.edu.cn.}
\thanks{D.~Wu and Y.~Tan are the corresponding authors.}}

\maketitle

\begin{abstract}
Takagi-Sugeno-Kang (TSK) fuzzy systems are very useful machine learning models for regression problems. However, to our knowledge, there has not existed an efficient and effective training algorithm that ensures their generalization performance, and also enables them to deal with big data. Inspired by the connections between TSK fuzzy systems and neural networks, we extend three powerful neural network optimization techniques, i.e., mini-batch gradient descent (MBGD), regularization, and AdaBound, to TSK fuzzy systems, and also propose three novel techniques (DropRule, DropMF, and DropMembership) specifically for training TSK fuzzy systems. Our final algorithm, MBGD with regularization, DropRule and AdaBound (MBGD-RDA), can achieve fast convergence in training TSK fuzzy systems, and also superior generalization performance in testing. It can be used for training TSK fuzzy systems on datasets of any size; however, it is particularly useful for big datasets, on which currently no other efficient training algorithms exist.
\end{abstract}


\IEEEpeerreviewmaketitle

\section{Introduction}

Fuzzy systems \cite{Mendel2017}, particularly Takagi-Sugeno-Kang (TSK) fuzzy systems \cite{Nguyen2019}, have achieved great success in numerous applications. This paper focuses on the applications of TSK fuzzy systems in machine learning \cite{Couso2019}, particularly, supervised regression problems. In such problems, we have a training dataset with $N$ labeled examples $\{\mathbf{x}_n,y_n\}_{n=1}^N$, where $\mathbf{x}_n\in\mathbb{R}^{M\times1}$, and would like to train a TSK fuzzy system to model the relationship between $y$ and $\mathbf{x}$, so that an accurate prediction can be made for any future unseen $\mathbf{x}$.

There are generally three different strategies for optimizing a TSK fuzzy system in supervised regression\footnote{Some novel approaches for optimizing evolving fuzzy systems have also been proposed recently \cite{Gu2019,Gu2019a}; however, they are not the focus of this paper, so their details are not included.}:
\begin{enumerate}
\item \emph{Evolutionary algorithms} \cite{drwuEAAI2006}, in which each set of the parameters of the antecedent membership functions (MFs) and the consequents are encoded as an individual in a population, and genetic operators, such as selection, crossover, mutation, and reproduction, are used to produce the next generation. Generally, the overall fitness improves in each new generation, and a global optimum may be found given enough number of generations.
\item \emph{Gradient descent} (GD) \cite{Wang1992b}, in which the parameters are moved in the negative gradient direction of the loss function to find its local minimum. Back-propagation  \cite{Rumelhart1986} is frequently used to calculate the gradients. These fuzzy systems are called \emph{neuro-fuzzy systems} in the literature \cite{Lin1996a}.
\item \emph{GD and least squares estimation (LSE)} \cite{Jang1993}, which is used in the popular adaptive-network-based fuzzy inference system (ANFIS). The antecedent parameters are optimized by GD, and the consequent parameters by LSE. This approach usually converges much faster than using GD only.
\end{enumerate}

However, all three strategies may have challenges in big data applications \cite{Chen2014a,Wu2014}. It's well-known that big data has at least three Vs\footnote{There may be also other Vs, e.g., veracity, value, etc.} \cite{McAfee2012}: volume (the size of the data), velocity (the speed of the data), and variety (the types of data). Volume means that the number of training examples ($N$) is very large, and/or the dimensionality of the input ($M$) is very high. Fuzzy systems, and actually almost all machine learning models, suffer from the curse of dimensionality, i.e., the number of rules (parameters) increases exponentially with $M$. However, in this paper we assume that the dimensionality can be reduced effectively to just a few, e.g., using principal component analysis \cite{Jolliffe2002}. We mainly consider how to deal with large $N$.

Evolutionary algorithms are not suitable for optimizing TSK fuzzy systems when $N$ is large, because they have very high memory and computing power requirement: they need to evaluate the fitness of each individual on the entire training dataset (which may be too large to be loaded into the memory completely), and there are usually tens or hundreds of individuals in a population, and tens or hundreds of generations are needed to find a good solution. ANFIS may result in significant overfitting in regression problems, as demonstrated in Section~\ref{sect:ANFIS} of this paper. So, we focus on GD.

When $N$ is small, batch GD can be used to compute the average gradients over all $N$ training examples, and then update the model parameters. When $N$ is large, there may not be enough memory to load the entire training dataset, and hence batch GD may be very slow or even impossible to perform. In such cases, stochastic GD can be used to compute the gradients for each training example, and then update the model parameters. However, the stochastic gradients may have very large variance, and hence the training may be unstable. A good compromise between batch GD and stochastic GD, which has achieved great success in deep learning \cite{Goodfellow2016}, is mini-batch gradient descent (MBGD). It randomly selects a small number (typically 32 or 64 \cite{Bengio2012}) of training examples to compute the gradients and update the model parameters. MBGD is a generic approach not specific to a particular model to be optimized, so it should also be applicable to the training of fuzzy systems. In fact, \cite{Nakasima-Lopez2019} has compared the performances of full-batch GD, MBGD and stochastic GD on the training of Mamdani neuro-fuzzy systems, and showed that MBGD achieved the best performance. This paper applies MBGD to the training of TSK fuzzy systems.

In MBGD, the learning rate is very important to the convergence speed and quality in training. Many different schemes, e.g., momentum \cite{Rumelhart1986}, averaging \cite{Polyak1992}, AdaGrad \cite{Duchi2011}, RMSProp \cite{Tieleman2012}, Adam \cite{Kingma2015}, etc., have been proposed to optimize the learning rate in neural network training. Adam may be the most popular one among them. However, to the knowledge of the authors, only a short conference paper \cite{Matsumura2017} has applied Adam to the training of simple single-input rule modules fuzzy systems \cite{Yubazaki1997}. Very recently, an improvement to Adam, AdaBound \cite{Luo2019}, was proposed, which demonstrated faster convergence and better generalization than Adam. To our knowledge, no one has used AdaBound for training TSK fuzzy systems.

In addition to fast convergence, the generalization ability of a machine learning model is also crucially important. Generalization means the model must perform well on previously unobserved inputs (not just the known training examples).

Regularization is frequently used to reduce overfitting and improve generalization. According to Goodfellow \emph{et al.} \cite{Goodfellow2016}, regularization is ``\emph{any modification we make to a learning algorithm that is intended to reduce its generalization error but not its training error.}" It has also been used in training TSK fuzzy systems to increase their performance and interpretability \cite{Johansen1996,Jin2000,Lughofer2010,Luo2015,Lughofer2008}. For example, Johansen \cite{Johansen1996}, and  Lughofer and Kindermann \cite{Lughofer2008}, used $\ell_2$ regularization (also known as weight decay, ridge regression, or Tikhonov regularization) to stabilize the matrix inversion operation in LSE. Jin \cite{Jin2000} used regularization to merge similar MFs into a single one to reduce the size of the rulebase and hence to increase the interpretability of the fuzzy system. Lughofer and Kindermann \cite{Lughofer2010}, and Luo \emph{et al.} \cite{Luo2015}, used sparsity regularization to identify a TSK fuzzy system with a minimal number of fuzzy rules and a minimal number of non-zero consequent parameters. All these approaches used LSE to optimize the TSK rule consequents, which may result in significant overfitting in regression problems (Section~\ref{sect:ANFIS}). To our knowledge, no one has integrated MBGD and regularization for TSK fuzzy system training.

Additionally, some unique approaches have also been proposed in the last few years to reduce overfitting and increase generalization of neural networks, particularly deep neural networks, e.g., DropOut \cite{Srivastava2014} and DropConnect \cite{Wan2013}. DropOut randomly discards some neurons and their connections during the training, which prevents units from co-adapting too much. DropConnect randomly sets some connection weights to zero during the training. Although DropOut and DropConnect have demonstrated outstanding performance and hence widely used in deep learning, no similar techniques exist for training TSK fuzzy systems.

This paper fills the gap in efficient and effective training of TSK fuzzy systems, particularly for big data regression problems. Its main contributions are:
\begin{enumerate}
\item Inspired by the connections between TSK fuzzy systems and neural networks \cite{drwuTSK2019}, we extend three powerful neural network optimization techniques, i.e., MBGD, regularization, and AdaBound, to TSK fuzzy systems.
\item We propose three novel techniques (DropRule, DropMF, and DropMembership) specifically for training TSK fuzzy systems.
\item Our final algorithm, MBGD with regularization, DropRule and AdaBound (MBGD-RDA), demonstrates superior performance on 10 real-world datasets from various application domains, of different sizes.
\end{enumerate}

The remainder of this paper is organized as follows: Section~\ref{sect:algorithm} introduces our proposed MBGD-RDA algorithm. Section~\ref{sect:experiment} presents our experimental results. Section~\ref{sect:conclusion} draws conclusion and points out some future research directions.

\section{The MBGD-RDA Algorithm} \label{sect:algorithm}

This section introduces our proposed MBGD-RDA algorithm for training TSK fuzzy systems, whose pseudo-code is given in Algorithm~\ref{alg:MBGD-RDA} and Matlab implementation at https://github.com/drwuHUST/MBGD\_RDA. Note that $\bm{\theta}_K$ returned from Algorithm~\ref{alg:MBGD-RDA} is not necessarily the optimal one among $\{\bm{\theta}_k\}_{k=1}^K$, i.e., the one that gives the smallest test error. The iteration number corresponding to the optimal $\bm{\theta}^*$ can be estimated using early stopping \cite{Goodfellow2016}. However, this is beyond the scope of this paper. Herein, we assume that the user has pre-determined $K$.

\begin{algorithm*}[htbp] %\DontPrintSemicolon
\KwIn{$N$ labeled training examples $\{\mathbf{x}_n,y_n\}_{n=1}^N$, where $\mathbf{x}_n=(x_{n,1},...,x_{n,M})^T\in \mathbb{R}^{M\times1}$\;
\hspace*{9mm} $L(\bm{\theta})$, the loss function for the TSK fuzzy system parameter vector $\bm{\theta}$\;
\hspace*{9mm} $M_m$, the number of Gaussian MFs in the $m$th input domain\;
\hspace*{9mm} $K$, the maximum number of training iterations\;
\hspace*{9mm} $N_{bs}\in[1,N]$, the mini-batch size\;
\hspace*{9mm} $P\in(0,1)$, the DropRule rate\;
\hspace*{9mm} $\alpha$, the initial learning rate (step size)\;
\hspace*{9mm} $\lambda$, the $\ell_2$ regularization coefficient\;
\hspace*{9mm} $\beta_1,\beta_2\in[0,1)$, exponential decay rates for the moment estimates in AdaBound\;
\hspace*{9mm} $\epsilon$, a small positive number in AdaBound\;
\hspace*{9mm} $l(k)$ and $u(k)$, the lower and upper bound functions in AdaBound\;}
\KwOut{The final $\bm{\theta}$.}
%\tcp{Initialization}
\For{$m=1,...,M$}
{Compute the minimum and maximum of all $\{x_{n,m}\}_{n=1}^N$\;
Initialize the center of the $M_m$ Gaussian MFs uniformly between the minimum and the maximum\;
Initialize the standard deviation of all $M_m$ Gaussian MFs as the standard deviation of $\{x_{n,m}\}_{n=1}^N$\;}
Initialize the consequent parameters of all $R$ rules as $0$\;
$\bm{\theta}_0$ is the concatenation of all Gaussian MF centers, standard deviations, and rule consequent parameters\;
%\tcp{Update $\bm{\theta}$}
$\mathbf{m}_0=\bm{0}$;
$\mathbf{v}_0=\bm{0}$\;
\For{$k=1,...,K$}
{%\tcp{MBGD}
Randomly select $N_{bs}$ training examples\;
\For{$n=1,...,N_{bs}$}{
\For{$r=1,...,R$}{
\tcp{DropRule}
$f_r(\mathbf{x}_n)=0$\;
Generate $p$, a uniformly distributed random number in $[0,1]$\;
\If{$p\le P$}{
Compute $f_r(\mathbf{x}_n)$, the firing level of $\mathbf{x}_n$ on $\mathrm{Rule}_r$\;}}
Compute $y(\mathbf{x}_n)$, the TSK fuzzy system output for $\mathbf{x}_n$, by (\ref{eq:yTSK})\;
\tcp{Compute the gradients}
\For{each element $\bm{\theta}_{k-1}(i)$ in $\bm{\theta}_{k-1}$}{
\begin{align*}
\mathbf{g}_k(i)=\left\{\begin{array}{ll}
                                       \frac{\partial L}{\partial \bm{\theta}_{k-1}(i)}, & \mbox{if }\bm{\theta}_{k-1}(i) \mbox{ was used in computing } y(\mathbf{x}_n) \\
                                       0, & \mbox{otherwise}
                                     \end{array}\right. &&&
\end{align*}}}
\tcp{$\ell_2$ regularization}
Identify the index set $I$, which consists of the elements of $\boldsymbol{\theta}$ corresponding to the rule consequent coefficients, excluding the bias terms\;
\For{each index $i\in I$}{
$\mathbf{g}_k(i)=\mathbf{g}_k(i)+\lambda\cdot \boldsymbol{\theta}_{k-1}(i)$\;}
\tcp{AdaBound}
$\mathbf{m}_k=\beta_1\bm{m}_{k-1}+(1-\beta_1)\bm{g}_k$;\quad
$\bm{v}_k=\beta_2\bm{v}_{k-1}+(1-\beta_2)\bm{g}_k^2$\;
$\displaystyle\hat{\bm{m}}_k=\frac{\bm{m}_k}{1-\beta_1^k}$; \quad
$\displaystyle\hat{\bm{v}}_k=\frac{\bm{v}_k}{1-\beta_2^k}$\;
$\displaystyle\hat{\bm{\alpha}}=\max\left[l(k),\min\left(u(k),
\frac{\alpha}{\sqrt{\hat{\bm{v}}_t}+\epsilon}\right)\right]$\;
$\bm{\theta}_k=\bm{\theta}_{k-1}-\hat{\bm{\alpha}}\odot\hat{\bm{m}}_k$\;
}
\textbf{Return} $\bm{\theta}_K$
\caption{The MBGD-RDA algorithm for TSK fuzzy system optimization. Typical values of some hyper-parameters are: $\beta_1=0.9$, $\beta_2=0.999$, and $\epsilon=10^{-8}$.} \label{alg:MBGD-RDA}
\end{algorithm*}

The key notations used in this paper are summarized in Table~\ref{tab:annotations}. The details of MBGD-RDA are explained next.

\begin{table*}[htbp] \centering  \setlength{\tabcolsep}{2mm}
\caption{Key notations used in this paper.}   \label{tab:annotations}
\begin{tabular}{c|l}   \hline
Notation    &\multicolumn{1}{|c}{Definition} \\ \hline
$N$ & The number of labeled training examples \\
$\mathbf{x}_n=(x_{n,1},...,x_{n,M})^T$ & $M$-dimension feature vector of the $n$th training example. $n\in[1,N]$ \\
$y_n$ & The groundtruth output corresponding to $\mathbf{x}_n$ \\
$R$ & The number of rules in the TSK fuzzy system \\
$X_{r,m}$ & The MF for the $m$th feature in the $r$th rule. $r\in[1,R]$, $m\in[1,M]$ \\
$b_{r,0},...,b_{r,M}$ & Consequent parameters of the $r$th rule. $r\in[1,R]$\\
$y_r(\mathbf{x}_n)$ & The output of the $r$th rule for $\mathbf{x}_n$. $r\in[1,R]$, $n\in[1,N]$\\
$\mu_{X_{r,m}}(x_{n,m})$ & The membership grade of $x_{n,m}$ on $X_{r,m}$. $r\in[1,R]$, $m\in[1,M]$, $n\in[1,N]$\\
$f_r(\mathbf{x}_n)$ & The firing level of $\mathbf{x}_n$ on the $r$th rule. $r\in[1,R]$, $n\in[1,N]$ \\
$y(\mathbf{x}_n)$ & The output of the TSK fuzzy system for $\mathbf{x}_n$ \\
$L$ & $\ell_2$ regularized loss function for training the TSK fuzzy system \\
$\lambda$   & The $\ell_2$ regularization coefficient in ridge regression, MBGD-R, MBGD-RA, and MBGD-RDA \\
$M_m$          & Number of Gaussian MFs in each input domain\\
$N_{bs}$        & Mini-batch size in MBGD-based algorithms\\
$K$       & Number of iterations in MBGD training \\
$\alpha$        & The initial learning rate in MBGD-based algorithms\\
$P$       & The DropRule rate in MBGD-D, MBGD-RD and MBGD-RDA\\
$\beta_1$, $\beta_2$ & The exponential decay rates for moment estimates in AdaBound\\
$\epsilon$ & A small positive number in AdaBound to avoid dividing by zero \\ \hline
\end{tabular}
\end{table*}

\subsection{The TSK Fuzzy System}

Assume the input $\mathbf{x}=(x_1,...,x_M)^T\in \mathbb{R}^{M\times1}$, and the TSK fuzzy system has $R$ rules:
\begin{align}
\mathrm{Rule}_r: &\mbox{ IF } x_1 \mbox{ is } X_{r,1} \mbox{ and } \cdots \mbox{ and } x_M \mbox{ is } X_{r,M}, \nonumber \\
&\mbox{ THEN } y_r(\mathbf{x})=b_{r,0}+\sum_{m=1}^M b_{r,m}x_m, \label{eq:Rr}
\end{align}
where $X_{r,m}$ ($r=1,...,R$; $m=1,...,M$) are fuzzy sets, and $b_{r,0}$ and $b_{r,m}$ are consequent parameters. This paper considers only Gaussian MFs, because their derivatives are easier to compute \cite{drwuMFs2012}. However, our algorithm can also be applied to other MF shapes, as long as their derivatives can be computed.

The membership grade of $x_{m}$ on a Gaussian MF $X_{r,m}$ is:
\begin{align}
\mu_{X_{r,m}}(x_{m})=\exp\left(-\frac{(x_{m}-c_{r,m})^2}{2\sigma_{r,m}^2}\right), \label{eq:mu}
\end{align}
where $c_{r,m}$ is the center of the Gaussian MF, and $\sigma_{r,m}$ the standard deviation.

The firing level of $\mathrm{Rule}_r$ is:
\begin{align}
f_r(\mathbf{x})=\prod_{m=1}^M\mu_{X_{r,m}}(x_{m}), \label{eq:fr}
\end{align}
and the output of the TSK fuzzy system is:
\begin{align}
y(\mathbf{x})=\frac{\sum_{r=1}^R f_r(\mathbf{x})y_r(\mathbf{x})}{\sum_{r=1}^R f_r(\mathbf{x})}. \label{eq:yTSK}
\end{align}
Or, if we define the normalized firing levels as:
\begin{align}
\bar{f}_r(\mathbf{x})=\frac{f_r(\mathbf{x})}{\sum_{k=1}^Rf_k(\mathbf{x})},\quad r=1,...,R \label{eq:f}
\end{align}
then, (\ref{eq:yTSK}) can be rewritten as:
\begin{align}
y(\mathbf{x})&=\sum_{r=1}^R\bar{f}_r(\mathbf{x})\cdot y_r(\mathbf{x}). \label{eq:yTSK2}
\end{align}

To optimize the TSK fuzzy system, we need to tune $c_{r,m}$, $\sigma_{r,m}$, $b_{r,0}$ and $b_{r,m}$, where $r=1,...,R$ and $m=1,...,M$.

\subsection{Regularization}

Assume there are $N$ training examples $\{\mathbf{x}_n,y_n\}_{n=1}^N$, where $\mathbf{x}_n=(x_{n,1},...,x_{n,M})^T\in \mathbb{R}^{M\times1}$.

In this paper, we use the following $\ell_2$ regularized loss function:
\begin{align}
L=\frac{1}{2}\sum_{n=1}^{N_{bs}}\left[y_n-y(\mathbf{x}_n)\right]^2+\frac{\lambda }{2}\sum_{r=1}^R\sum_{m=1}^M b_{r,m}^2, \label{eq:loss}
\end{align}
where $N_{bs}\in[1,N]$, and $\lambda\ge 0$ is a regularization parameter. Note that $b_{r,0}$ ($r=1,...,R$) are not regularized in (\ref{eq:loss}). As pointed out by Goodfellow \emph{et al.} \cite{Goodfellow2016}, for neural networks, one typically penalizes only the weights of the affine transformation at each layer and leaves the biases un-regularized. The biases typically require less data to fit accurately than the weights. The biases in neural networks are corresponding to the $b_{r,0}$ terms here, so we take this typical approach, and leave $b_{r,0}$ un-regularized.

\subsection{Mini-Batch Gradient Descent (MBGD)}

The gradients of the loss function (\ref{eq:loss}) are given in (\ref{eq:dEodC})-(\ref{eq:dEdB}), where $\Phi(r,m)$ is the index set of the rules that contain $X_{r,m}$, $x_{n,0}\equiv1$, and $I(m)$ is an indicator function:
\begin{align*}
I(m)=\left\{\begin{array}{cc}
              0, & m=0 \\
              1, & m>0
            \end{array}\right.
\end{align*}
$I(m)$ ensures that $b_{r,0}$ ($r=1,...,R$) are not regularized.

\begin{figure*}
\begin{align}
\frac{\partial L}{\partial c_{r,m}}&=\frac{1}{2}\sum_{n=1}^{N_{bs}}\sum_{k=1}^R\frac{\partial L}{\partial y(\mathbf{x}_n)}
\frac{\partial y(\mathbf{x}_n)}{\partial f_k(\mathbf{x}_n)}
\frac{\partial f_k(\mathbf{x}_n)}{\partial \mu_{X_{k,m}}(x_{n,m})}
\frac{\partial \mu_{X_{k,m}}(x_{n,m})}{\partial c_{r,m}}\nonumber\\
&=\sum_{n=1}^{N_{bs}}\sum_{k\in \Phi(r,m)}\left[\left(y(\mathbf{x}_n)-y_n\right)
\frac{y_k(\mathbf{x}_n)\sum_{i=1}^Rf_i(\mathbf{x}_n)
-\sum_{i=1}^Rf_i(\mathbf{x}_n)y_i(\mathbf{x}_n)}
{\left[\sum_{i=1}^Rf_i(\mathbf{x}_n)\right]^2}f_k(\mathbf{x}_n)
\frac{x_{n,m}-c_{r,m}}{\sigma_{r,m}^2}\right] \label{eq:dEodC}\\
\frac{\partial L}{\partial \sigma_{r,m}}&=\frac{1}{2}\sum_{n=1}^{N_{bs}}\sum_{k=1}^R\frac{\partial L}{\partial y(\mathbf{x}_n)}
\frac{\partial y(\mathbf{x}_n)}{\partial f_k(\mathbf{x}_n)}
\frac{\partial f_k(\mathbf{x}_n)}{\partial \mu_{X_{k,m}}(x_{n,m})}
\frac{\partial \mu_{X_{k,m}}(x_{n,m})}{\partial \sigma_{r,m}}\nonumber\\
&=\sum_{n=1}^{N_{bs}}\sum_{k\in\Phi(r,m)}\left[\left(y(\mathbf{x}_n)-y_n\right)
\frac{y_k(\mathbf{x}_n)\sum_{i=1}^Rf_i(\mathbf{x}_n)
-\sum_{i=1}^Rf_i(\mathbf{x}_n)y_i(\mathbf{x}_n)}
{\left[\sum_{i=1}^Rf_i(\mathbf{x}_n)\right]^2}f_k(\mathbf{x}_n)
\frac{(x_{n,m}-c_{r,m})^2}{\sigma_{r,m}^3}\right] \label{eq:dEodSigma}\\
\frac{\partial L}{\partial b_{r,m}}&=\frac{1}{2}\sum_{n=1}^{N_{bs}}\frac{\partial L}{\partial y(\mathbf{x}_n)}
\frac{\partial y(\mathbf{x}_n)}{\partial y_r(\mathbf{x}_n)}
\frac{\partial y_r(\mathbf{x}_n)}{\partial b_{r,m}}+\frac{\lambda}{2} \frac{\partial L}{\partial b_{r,m}}=\sum_{n=1}^{N_{bs}}\left[\left(y(\mathbf{x}_n)-y_n\right) \frac{f_r(\mathbf{x}_n)}{\sum_{i=1}^Rf_i(\mathbf{x}_n)}\cdot x_{n,m}\right] +\lambda I(m)b_{r,m}\label{eq:dEdB}
\end{align}
\end{figure*}

In MBGD, each time we randomly sample $N_{bs}\in[1, N]$ training examples, compute the gradients from them, and then update the antecedent and consequent parameters of the TSK fuzzy system. Let $\bm{\theta}_k$ be the model parameter vector in the $k$th iteration, and $\partial{L}/\partial{\bm{\theta}_k}$ be the first-order gradients computed from  (\ref{eq:dEodC})-(\ref{eq:dEdB}). Then, the update rule is:
\begin{align}
\bm{\theta}_k=\bm{\theta}_{k-1}-\alpha\frac{\partial{L}}{\partial{\bm{\theta}_{k-1}}},
\end{align}
where $\alpha>0$ is the learning rate (step size).

When $N_{bs}=1$, MBGD degrades to the stochastic GD. When $N_{bs}=N$, it becomes the batch GD.

\subsection{DropRule}

DropRule is a novel technique to reduce overfitting and increase generalization in training TSK fuzzy systems, inspired by the well-known DropOut \cite{Srivastava2014} and DropConnect \cite{Wan2013} techniques in deep learning. DropOut randomly discards some neurons and their connections during the training. DropConnect randomly sets some connection weights to zero during the training. DropRule randomly discards some rules during the training, but uses all rules in testing.

\begin{figure}[htpb] \centering
\subfigure[]{\label{fig:dropRule1}     \includegraphics[width=.9\linewidth,clip]{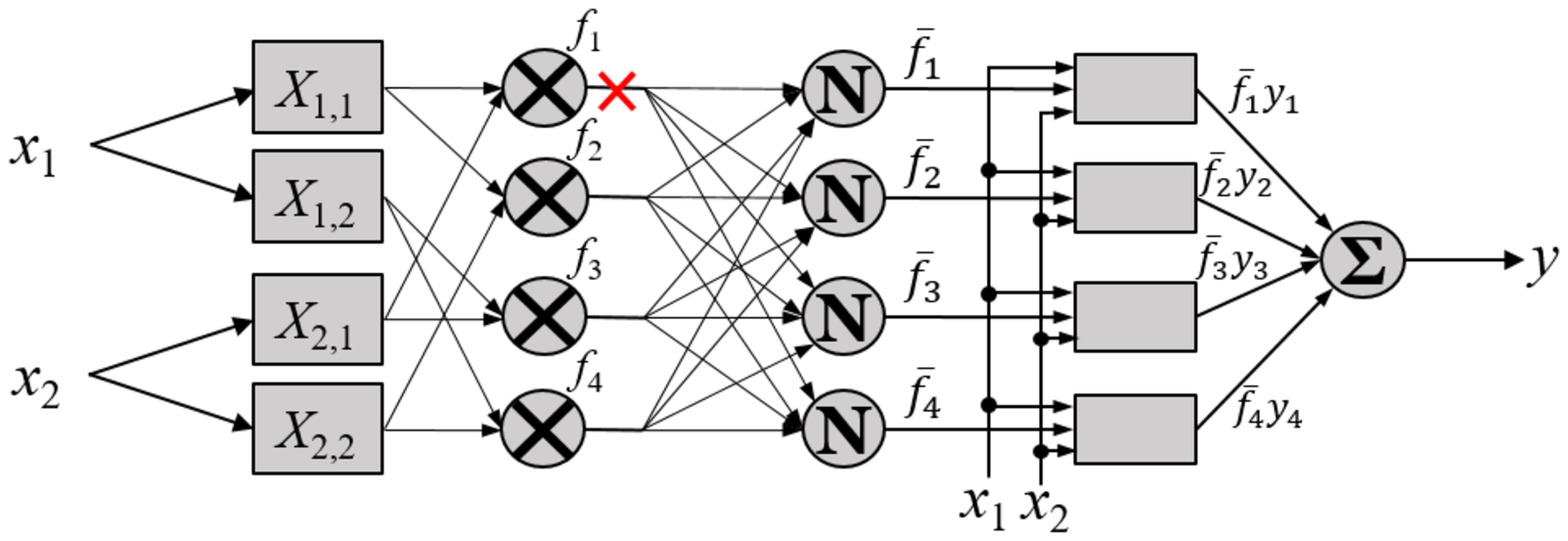}}
\subfigure[]{\label{fig:dropRule2}     \includegraphics[width=.9\linewidth,clip]{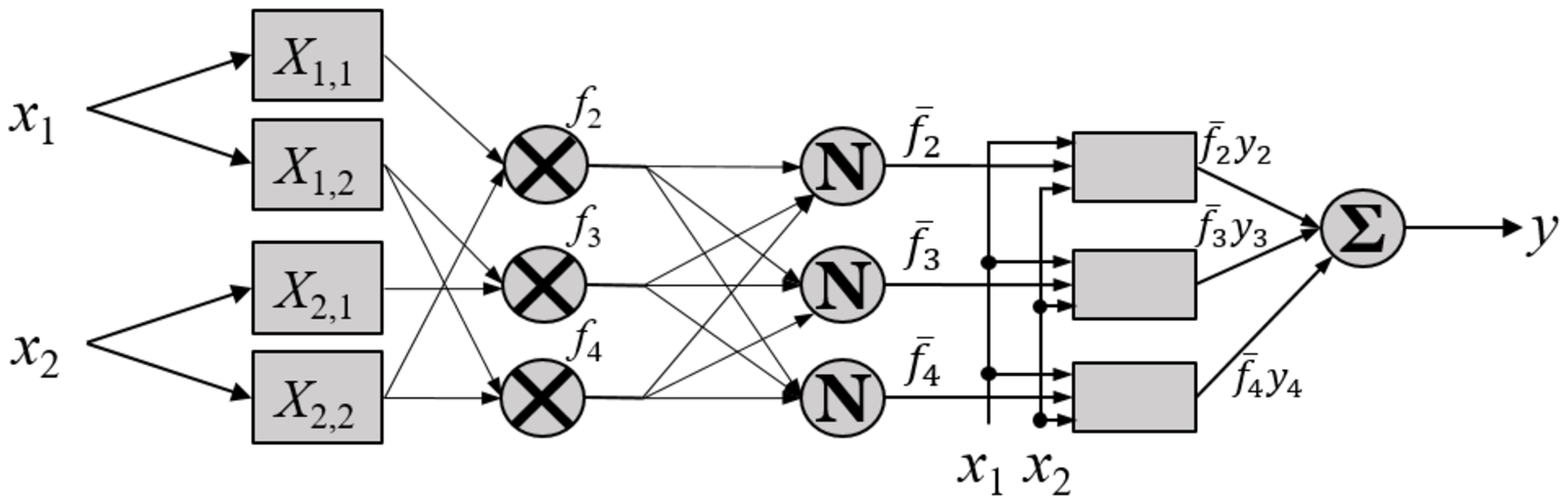}}
\caption{DropRule, where $X_{m,i}$ is the $i$th MF in the $m$th input domain. (a) The red cross indicates that the first rule will be dropped; and, (b) the equivalent fuzzy system after dropping the first rule.} \label{fig:dropRule}
\end{figure}

Let the DropRule rate be $P\in(0,1)$. For each training example in the iteration, we set the firing level of a rule to its true firing level with probability $P$, and to zero with probability $1-P$. The output of the TSK fuzzy system is again computed by a firing level weighted average of the rule consequents. Since the firing levels of certain rules are artificially set to zero, they do not contribute anything to the fuzzy system output, i.e., they are artificially dropped for this particular training example, as shown in Fig.~\ref{fig:dropRule}\footnote{The ANFIS representation of a TSK fuzzy system is used here. For details, please refer to Section~\ref{sect:ANFIS}.}. Then, GD is applied to update the model parameters in rules that are not dropped (the parameters in the dropped rules are not updated for this particular training example).

When the training is done, all rules will be used in computing the output for a new input, just as in a traditional TSK fuzzy system. This is different from DropOut and DropConnect for neural networks, which need some special operations in testing to ensure the output does not have a bias. We do not need to pay special attention in using a TSK fuzzy system trained from DropRule, because the final step of a TSK fuzzy system is a weighted average, which removes the bias automatically.

The rationale behind DropOut is that \cite{Srivastava2014} \emph{``each hidden unit in a neural network trained with dropout must learn to work with a randomly chosen sample of other units. This should make each hidden unit more robust and drive it towards creating useful features on its own without relying on other hidden units to correct its mistakes."} This is also the motivation of DropRule: by randomly dropping some rules, we force each rule to work with a \emph{randomly} chosen subset of rules, and hence each rule should maximize its own modeling capability, instead of relying too much on other rules. This may help increase the generalization of the final TSK fuzzy system. Our experiments in the next section demonstrate that DropRule alone may not always offer advantages, but it works well when integrated with an efficient learning rate adaptation algorithm like AdaBound.

\subsection{Adam and AdaBound}

As mentioned in the Introduction, the learning rate is a very important hyper-parameter in neural network training, which is also the case for TSK fuzzy system training. Among the various proposals for adjusting the learning rate, Adam \cite{Kingma2015} may be the most popular one. It computes an individualized adaptive learning rate for each different model parameter from the estimates of the first and second moments of the gradient. Essentially, it combines the advantages of two other approaches: AdaGrad \cite{Duchi2011}, which works well with sparse gradients, and RMSProp \cite{Tieleman2012}, which works well in online and non-stationary settings.

Very recently, an improvement to Adam, AdaBound \cite{Luo2019}, was proposed. It bounds the individualized adaptive learning rate from the upper and the lower, so that an extremely large or small learning rate cannot occur. Additionally, the bounds become tighter as the number of iterations increases, which forces the learning rate to approach a constant (as in the stochastic GD). AdaBound has demonstrated faster convergence and better generalization than Adam in \cite{Luo2019}, so it is adopted in this paper.

The pseudo-code of AdaBound can be found in Algorithm~\ref{alg:MBGD-RDA}. By substituting $L$ in (\ref{eq:loss}) into it, we can use the bounded individualized adaptive learning rates for different elements of $\bm{\theta}$, which may result in better training and generalization performance than using a fixed learning rate. The lower and upper bound functions used in this paper were similar to those in \cite{Luo2019}:
\begin{align}
l(k)&=0.01-\frac{0.01}{(1-\beta_2)k+1} \label{eq:lk}\\
u(k)&=0.01+\frac{0.01}{(1-\beta_2)k} \label{eq:uk}
\end{align}
When $k=0$, the bound is $[0,+\infty)$. When $k$ approaches $+\infty$, the bound approaches $[0.01,0.01]$.

\section{Experiments} \label{sect:experiment}

This section presents experimental results to demonstrate the performance of our proposed MBGD-RDA.

\subsection{Datasets}

Ten regression datasets from the CMU StatLib Datasets Archive\footnote{http://lib.stat.cmu.edu/datasets/} and the UCI Machine Learning Repository\footnote{http://archive.ics.uci.edu/ml/index.php} were used in our experiments. Their summary is given in Table~\ref{tab:Datasets}. Their sizes range from small to large.

\begin{table}[htbp] \centering  \setlength{\tabcolsep}{.8mm}
\caption{Summary of the 10 regression datasets.}   \label{tab:Datasets}
\begin{tabular}{c|cccccc}   \hline
Dataset    &Source &\tabincell{c}{No. of\\examples} &\tabincell{c}{No. of\\raw\\features} &\tabincell{c}{No. of\\numerical\\features}  &\tabincell{c}{No. of\\used\\features}  &\tabincell{c}{No. of TSK\\model\\parameters}  \\ \hline
PM10$^1$           &StatLib &500            &7            &7            &5 & 212\\
NO2$^1$            &StatLib &500            &7            &7            &5 & 212\\
Housing$^2$        &UCI     &506            &13           &13          &5 & 212\\
Concrete$^3$       &UCI     &1,030           &8            &8      &5 & 212\\
Airfoil$^4$        &UCI     &1,503           &5            &5         &5 & 212\\
Wine-Red$^5$       &UCI     &1,599           &11           &11        &5 & 212\\
Abalone$^6$        &UCI     &4,177           &8            &7         &5 & 212\\
Wine-White$^5$     &UCI     &4,898           &11           &11         &5 & 212\\
PowerPlant$^7$     &UCI     &9,568           &4            &4         &4 & 96\\
Protein$^8$        &UCI     &45,730          &9            &9         &5 &212\\ \hline
\end{tabular}\\ \raggedright
\footnotesize{$^1$ http://lib.stat.cmu.edu/datasets/}\\
\footnotesize{$^2$ https://archive.ics.uci.edu/ml/machine-learning-databases/housing/}\\
\footnotesize{$^3$ https://archive.ics.uci.edu/ml/datasets/Concrete+Compressive+Strength}\\
\footnotesize{$^4$ https://archive.ics.uci.edu/ml/datasets/Airfoil+Self-Noise}\\
\footnotesize{$^5$ https://archive.ics.uci.edu/ml/datasets/Wine+Quality}\\
\footnotesize{$^6$ https://archive.ics.uci.edu/ml/datasets/Abalone}\\
\footnotesize{$^7$ https://archive.ics.uci.edu/ml/datasets/Combined+Cycle+Power+Plant}\\
\footnotesize{$^8$ https://archive.ics.uci.edu/ml/datasets/Physicochemical+Properties+of+\\Protein+Tertiary+Structure}
\end{table}

Nine of the 10 datasets have only numerical features. Abalone has a categorical feature (sex: male, female, and infant), which was ignored in our experiments\footnote{We also tried to convert the categorical feature into numerical ones using one-hot coding and use them together with the other seven numerical features. However, the RMSEs were worse than simply ignoring it.}. Each numerical feature was $z$-normalized to have zero mean and unit variance, and the output mean was also subtracted. Because fuzzy systems have difficulty dealing with high dimensional data, we constrained the maximum input dimensionality to be five: if a dataset had more than five features, then principal component analysis was used to reduce them to five.

The TSK fuzzy systems had $M_m=2$ Gaussian MFs in each input domain. For $M$ inputs, the TSK fuzzy system has $2MM_m+(M+1)\cdot(M_m)^M$ parameters.

\subsection{Algorithms}

We compared the performances of the following seven algorithms\footnote{We also tried to use support vector regression \cite{Smola2004} as a baseline regression model; however, it was too time-consuming to train on big datasets. So, we abandoned it.}:
\begin{enumerate}
\item \emph{Ridge regression} \cite{Hoerl1970}, with ridge coefficient $\lambda=0.05$. This algorithm is denoted as \texttt{RR} in the sequel.

\item \emph{MBGD}, which is a mini-batch version of the batch GD algorithm introduced in \cite{Jang1993}. The batch size $N_{bs}$ was 64, the initial learning rate $\alpha$ was $0.01$, and the adaptive learning rate adjustment rule in \cite{Jang1993} was implemented: $\alpha$ was multiplied by $1.1$ if the loss function was reduced in four successive iterations, and by $0.9$ if the loss function had two consecutive combinations of an increase followed by a decrease. This algorithm is denoted as \texttt{MBGD} in the sequel.

\item \emph{MBGD with Regularization}, which was essentially identical to \texttt{MBGD}, except that the loss function had an $\ell_2$ regularization term on the consequent parameters [see (\ref{eq:loss})]. $\lambda=0.05$ was used. This algorithm is denoted as \texttt{MBGD-R} in the sequel.

\item \emph{MBGD with DropRule}, which was essentially identical to \texttt{MBGD}, except that DropRule with $P=0.5$ was also used in the training, i.e., for each training example, we randomly set the firing level of 50\% rules to zero. This algorithm is denoted as \texttt{MBGD-D} in the sequel.

\item \emph{MBGD with Regularization and DropRule}, which integrated \texttt{MBGD-R} and \texttt{MBGD-D}. This algorithm is denoted as \texttt{MBGD-RD} in the sequel.

\item \emph{MBGD with AdaBound}, which was essentially identical to \texttt{MBGD}, except that the learning rates were adjusted by AdaBound. $\alpha=0.01$, $\beta_1=0.9$, $\beta_2=0.999$, and $\epsilon=10^{-8}$ were used. This algorithm is denoted as \texttt{MBGD-A} in the sequel.

\item \emph{MBGD with Regularization, DropRule and AdaBound}, which combined \texttt{MBGD-R}, \texttt{MBGD-D} and \texttt{MBGD-A}. This algorithm is denoted as \texttt{MBGD-RDA} in the sequel.
\end{enumerate}
For clarity, the parameters of these seven algorithms are also summarized in Table~\ref{tab:parameters}.

\begin{table}[htbp] \centering  \setlength{\tabcolsep}{2mm}
\caption{Parameters of the seven algorithms used in the experiments. The definitions of the parameters can be found in Table~\ref{tab:annotations}.}   \label{tab:parameters}
\begin{tabular}{c|l}   \hline
Algorithm    &\multicolumn{1}{|c}{Parameters} \\ \hline
\texttt{RR}           & $\lambda=0.05$ \\
\texttt{MBGD}           & $M_m=2$, $N_{bs}=64$, $K=500$, $\alpha=0.01$\\
\texttt{MBGD-R}        & $M_m=2$, $N_{bs}=64$, $K=500$, $\alpha=0.01$, $\lambda=0.05$\\
\texttt{MBGD-D}       & $M_m=2$, $N_{bs}=64$, $K=500$, $\alpha=0.01$, $P=0.5$ \\
\texttt{MBGD-RD}        & $M_m=2$, $N_{bs}=64$, $K=500$, $\alpha=0.01$, $\lambda=0.05$,\\
& $P=0.5$\\
\texttt{MBGD-A}       & $M_m=2$, $N_{bs}=64$, $K=500$, $\alpha=0.01$,\\
 & $\beta_1=0.9$, $\beta_2=0.999$, $\epsilon=10^{-8}$\\
\texttt{MBGD-RDA}        & $M_m=2$, $N_{bs}=64$, $K=500$, $\alpha=0.01$, $\lambda=0.05$, \\
& $P=0.5$, $\beta_1=0.9$, $\beta_2=0.999$, $\epsilon=10^{-8}$ \\ \hline
\end{tabular}
\end{table}

For each dataset, we randomly selected 70\% examples for training, and the remaining 30\% for test. \texttt{RR} was trained in one single pass on all training examples, and then its root mean squared error (RMSE) on the test examples was computed. The other six MBGD-based algorithms were iterative. The TSK fuzzy systems had two Gaussian MFs in each input domain. Their centers were initialized at the minimum and maximum of the input domain, and their standard deviations were initialized to the standard deviation of the corresponding input. All rule consequent parameters were initialized to zero. The maximum number of iterations was 500. After each training iteration, we recorded the corresponding test RMSE of each algorithm. Because there was randomness involved (e.g., the training/test data partition, the selection of the mini-batches, etc.), each algorithm was repeated 10 times on each dataset, and the average test results are reported next.

\subsection{Experimental Results}

\begin{figure*}[htbp]\centering
\includegraphics[width=\linewidth,clip]{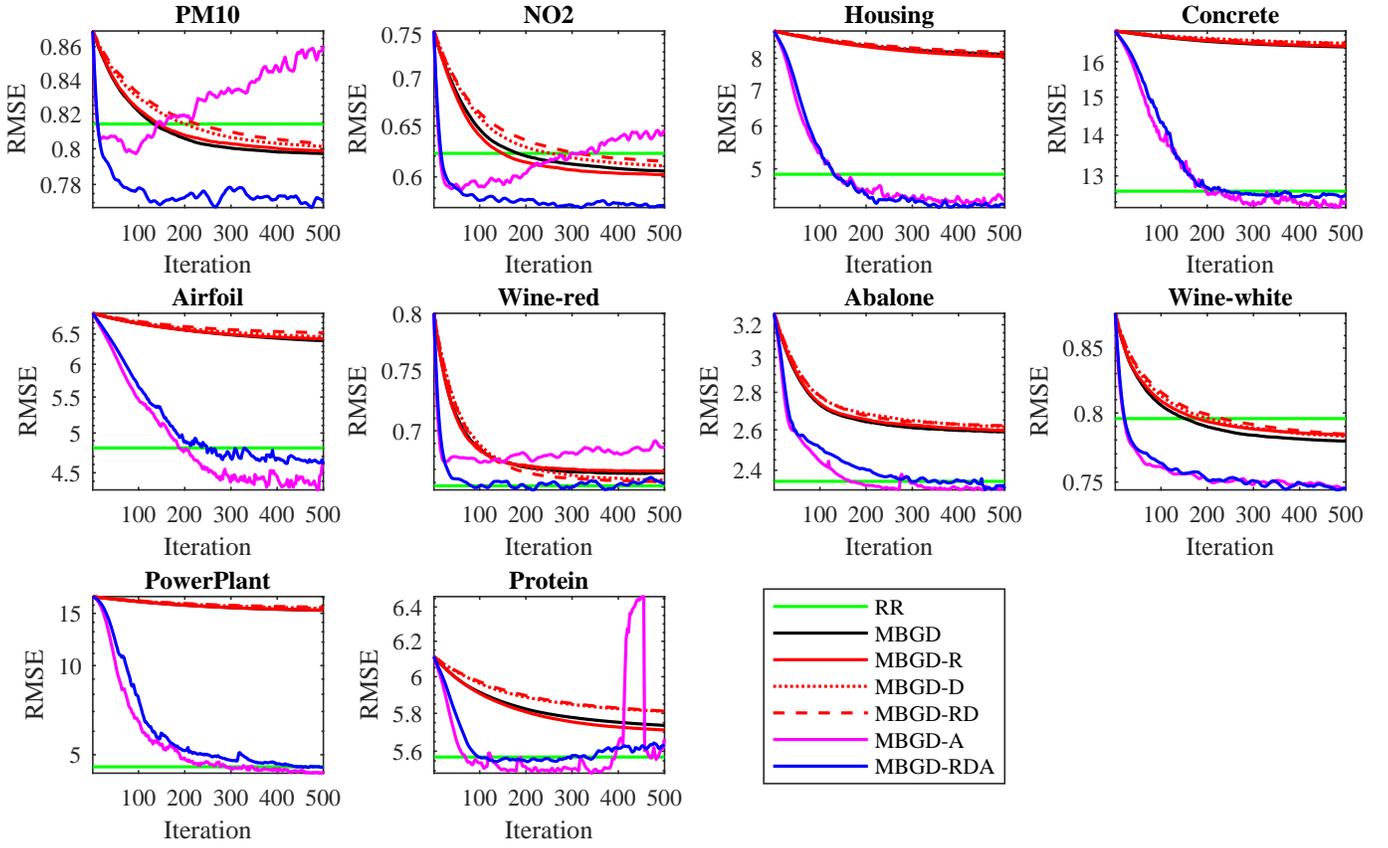}
\caption{The average test RMSEs of the seven algorithms on the 10 datasets.} \label{fig:mean}
\end{figure*}

The average test RMSEs of the seven algorithms are shown in Fig.~\ref{fig:mean}. We can observe that:
\begin{enumerate}
\item \texttt{MBGD-R}, \texttt{MBGD-D} and \texttt{MBGD-RD} had comparable performance with \texttt{MBGD}. All of them were worse than the simple \texttt{RR} on seven out of the 10 datasets, suggesting that a model with much more parameters and nonlinearity does not necessarily outperform a simple linear regression model, if not properly trained.

\item \texttt{MBGD-RDA} and \texttt{MBGD-A} performed the best among the seven algorithms. On nine out of the 10 datasets (except Wine-Red), \texttt{MBGD-A}'s best test RMSEs were smaller than \texttt{RR}. On all 10 datasets, \texttt{MBGD-RDA}'s best test RMSEs were smaller than \texttt{RR}. \texttt{MBGD-RDA} and \texttt{MBGD-A} also converged much faster than \texttt{MBGD}, \texttt{MBGD-R}, \texttt{MBGD-D} and \texttt{MBGD-RD}. As the final TSK fuzzy systems trained from the six MBGD-based algorithms had the same structure and the same number of parameters, these results suggest that AdaBound was indeed very effective in updating the learning rates, which in turn helped obtain better learning performances.

\item Although regularization alone (\texttt{MBGD-R}), DropRule alone (\texttt{MBGD-D}), and the combination of regularization and DropRule (\texttt{MBGD-RD}), did not result in much performance improvement (i.e., \texttt{MBGD-R}, \texttt{MBGD-D}, \texttt{MBGD-RD} and  \texttt{MBGD} had similar performances), \texttt{MBGD-RDA} outperformed \texttt{MBGD-A} on three out of the 10 datasets, and they had comparable performances on many other datasets. These results suggest that using an effective learning rate updating scheme like AdaBound can help unleash the power of regularization and DropRule, and hence achieve better learning performance.
\end{enumerate}

\begin{figure*}[htpb]\centering
\includegraphics[width=\linewidth,clip]{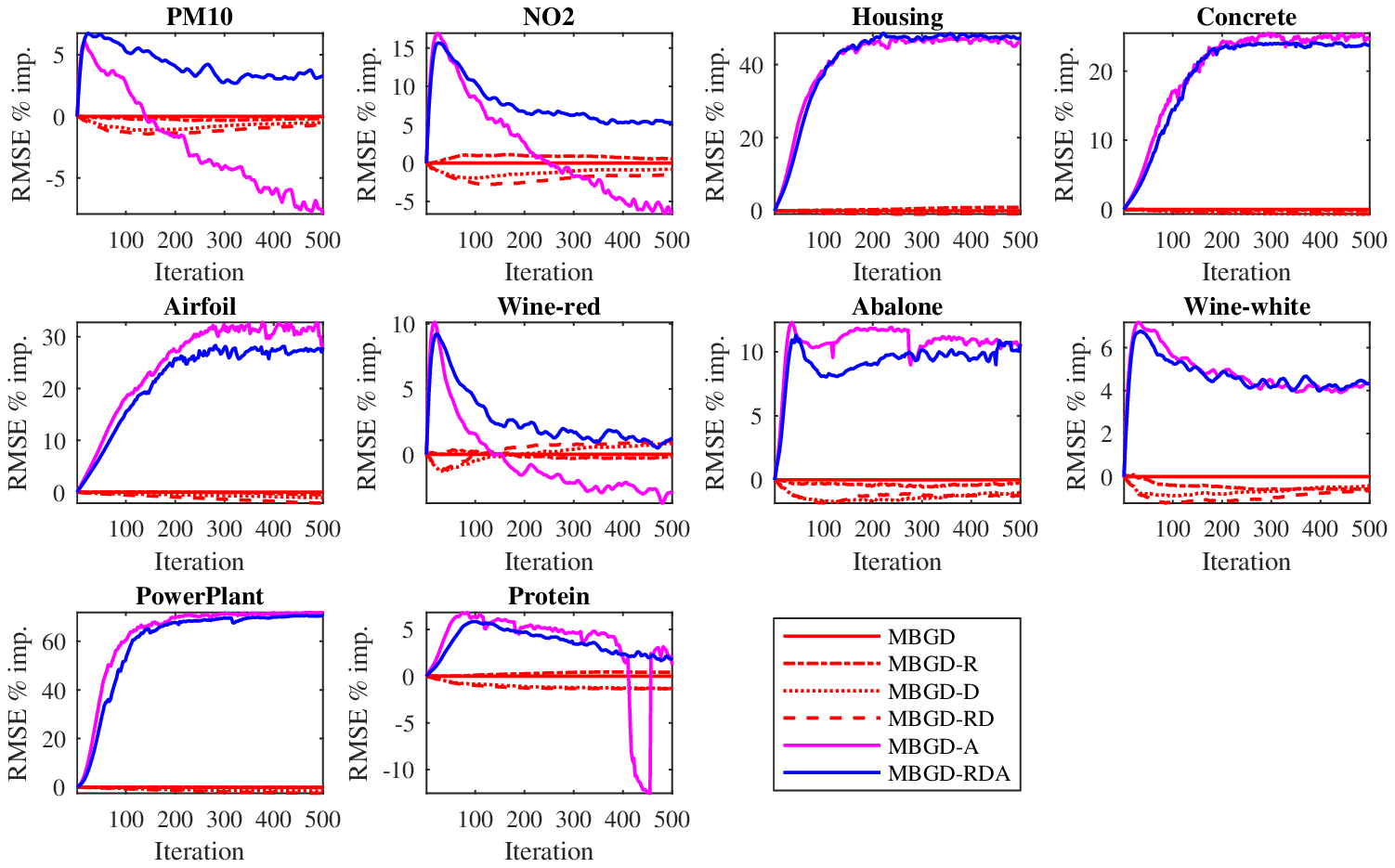}
\caption{Percentage improvements of the test RMSEs of \texttt{MBGD-R}, \texttt{MBGD-D}, \texttt{MBGD-RD}, \texttt{MBGD-A} and \texttt{MBGD-RDA} over \texttt{MBGD}.} \label{fig:imp}
\end{figure*}

To better visualize the performance differences among the six MBGD-based algorithms, we plot in Fig.~\ref{fig:imp} the percentage improvements of \texttt{MBGD-R}, \texttt{MBGD-D}, \texttt{MBGD-RD}, \texttt{MBGD-A} and \texttt{MBGD-RDA} over \texttt{MBGD}: in each iteration, we treat the test RMSE of \texttt{MBGD} as one, and compute the relative percentage improvements of the test RMSEs of the other five MBGD-based algorithms over it. For example, let $RMSE_{GD}(k)$ and $RMSE_{MBGD-RDA}(k)$ be the test RMSEs of \texttt{MBGD} and \texttt{MBGD-RDA} at iteration $k$, respectively. Then, the percentage improvement of the test RMSE of \texttt{MBGD-RDA} over \texttt{MBGD} at iteration $k$ is:
\begin{align}
p(k)=100\times\frac{RMSE_{GD}(k)-RMSE_{MBGD-RDA}(k)}{RMSE_{GD}(k)}.
\end{align}

Fig.~\ref{fig:imp} confirmed the observations made from Fig.~\ref{fig:mean}. Particularly, \texttt{MBGD-RDA} and \texttt{MBGD-A} converged much faster and to smaller values than \texttt{MBGD}, \texttt{MBGD-R}, \texttt{MBGD-D} and \texttt{MBGD-RD}; the best test RMSEs of \texttt{MBGD-RDA} and \texttt{MBGD-A} were also much smaller than those of \texttt{MBGD}, \texttt{MBGD-R}, \texttt{MBGD-D} and \texttt{MBGD-RD}. Among the five enhancements to \texttt{MBGD}, only \texttt{MBGD-RDA} consistently outperformed \texttt{MBGD}. In other words, although \texttt{MBGD-RDA} may not always outperform \texttt{MBGD-A}, its performance was more stable than \texttt{MBGD-A}; so, it should be preferred over \texttt{MBGD-A} in practice.

The time taken to finish 500 training iterations for the MBGD-based algorithms on the 10 datasets is shown in Table~\ref{tab:CC}. The platform was a desktop computer running Matlab 2018a and Windows 10 Enterprise 64x, with Intel Core i7-8700K CPU @ 3.70 GHz, 16GB memory, and 512GB solid state drive. The CPU has 12 cores, but each algorithm used only one core. Not surprisingly, \texttt{RR} was much faster than others, because it has a closed-form solution, and no iteration was needed. Among the six MBGD-based algorithms, \texttt{MBGD-RDA} was the fastest. One reason is that DropRule reduced the number of parameters to be adjusted in each iteration.

\begin{table}[htbp] \centering  \setlength{\tabcolsep}{.9mm}
\caption{Computational cost (seconds) of different algorithms on the 10 regression datasets.}   \label{tab:CC}
\begin{tabular}{c|ccccccc}   \hline
Dataset    &\texttt{RR} &\texttt{MBGD} &\tabincell{c}{\texttt{MBGD}\\ \texttt{-R}}  &\tabincell{c}{\texttt{MBGD}\\ \texttt{-D}}  &\tabincell{c}{\texttt{MBGD}\\ \texttt{-RD}}
&\tabincell{c}{\texttt{MBGD}\\ \texttt{-A}} &\tabincell{c}{\texttt{MBGD}\\ \texttt{-RDA}} \\ \hline
PM10           & 0.003&21.115&21.027&19.194&19.388&20.746&15.394\\
NO2            & 0.003&21.619&21.273&19.627&19.620&21.063&15.971\\
Housing        & 0.003&21.304&21.064&19.392&19.357&20.799&15.782\\
Concrete       & 0.003&29.891&30.143&27.784&27.943&30.142&25.468\\
Airfoil        & 0.005&35.813&35.928&33.800&34.532&36.608&31.027\\
Wine-Red       & 0.003&36.704&37.016&35.606&35.594&36.793&32.070\\
Abalone        & 0.003&38.780&39.046&38.406&38.861&38.909&36.921\\
Wine-White     & 0.003&74.429&75.992&73.091&74.525&74.844&70.531\\
PowerPlant     & 0.005&68.954&66.656&68.187&65.763&66.849&64.811\\
Protein        & 0.008&474.995&445.170&469.929&433.486&461.251&429.679\\ \hline
\end{tabular}
\end{table}

\subsection{Parameter Sensitivity}

It's also important to study the sensitivity of \texttt{MBGD-RDA} to its hyper-parameters. Algorithm~\ref{alg:MBGD-RDA} shows that \texttt{MBGD-RDA} has the following hyper-parameters:
\begin{enumerate}
\item $M_m$, the number of Gaussian MFs in the $m$th input domain
\item $K$, the maximum number of training iterations
\item $N_{bs}\in[1,N]$, the mini-batch size
\item $P\in(0,1)$, the DropRule rate
\item $\alpha$, the initial learning rate (step size)
\item $\lambda$, the $\ell_2$ regularization coefficient
\item $\beta_1,\beta_2\in[0,1)$, exponential decay rates for the moment estimates in AdaBound
\item $\epsilon$, a small positive number in AdaBound
\item $l(k)$ and $u(k)$, the lower and upper bound functions in AdaBound
\end{enumerate}
Among them, $M_m$ is a parameter for all TSK fuzzy systems, not specific to \texttt{MBGD-RDA}; $K$ can be determined by early-stopping on a validation dataset; and, $\beta_1$, $\beta_2$, $\epsilon$, $l(k)$ and $u(k)$ are AdaBound parameters, whose default values are usually used. So, we only studied the sensitivity of \texttt{MBGD-RDA} to $N_{bs}$, $P$, $\alpha$ and $\lambda$, which are unique to \texttt{MBGD-RDA}.

\begin{figure}[htpb] \centering
\subfigure[]{\label{fig:batchSize}     \includegraphics[width=.48\linewidth,clip]{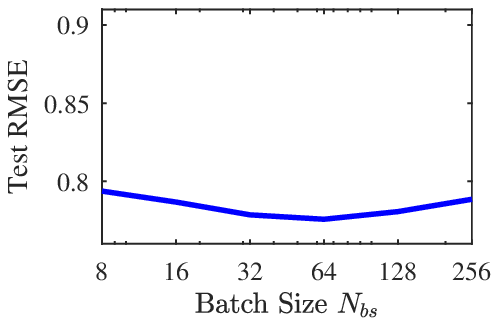}}
\subfigure[]{\label{fig:P}     \includegraphics[width=.48\linewidth,clip]{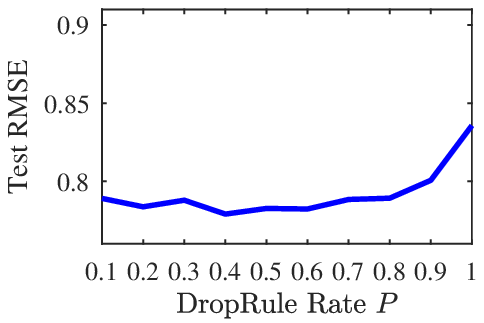}}
\subfigure[]{\label{fig:alpha}     \includegraphics[width=.48\linewidth,clip]{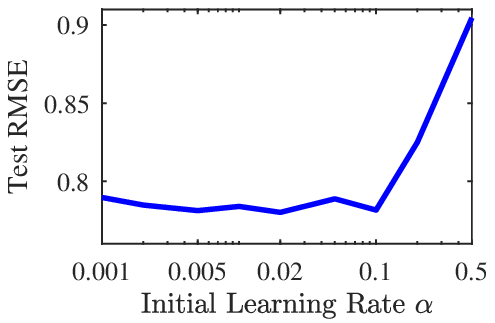}}
\subfigure[]{\label{fig:lambda}     \includegraphics[width=.48\linewidth,clip]{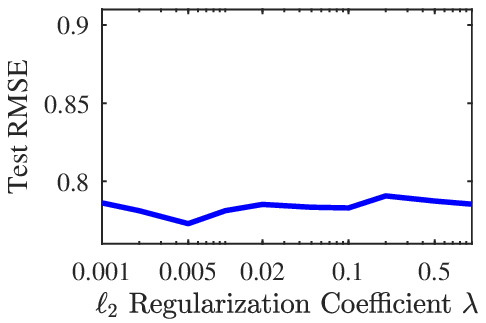}}
\caption{Test RMSEs of \texttt{MBGD-RDA} w.r.t. different hyper-parameters on the PM10 dataset. (a) Different batch size $N_{bs}$;  (b) different DropRule rate $P$; (c) different initial learning rate $\alpha$; and, (d) different $\ell_2$ regularization coefficient $\lambda$. In each subfigure, except for the hyper-parameter under consideration, the values for other parameters were: $M_m=2$, $K=500$, $N_{bs}=64$, $P=0.5$, $\alpha=0.01$, $\lambda=0.05$, $\beta_1=0.9$, $\beta_2=0.999$, $\epsilon=10^{-8}$, and $l(k)$ and $u(k)$ are defined in (\ref{eq:lk}) and (\ref{eq:uk}), respectively.} \label{fig:paraSens}
\end{figure}

The results, in terms of the test RMSEs, on the PM10 dataset are shown in Fig.~\ref{fig:paraSens}, where each experiment was repeated 100 times and the average test RMSEs are shown. In each subfigure, except for the hyper-parameter under consideration, the values for other parameters were: $M_m=2$, $K=500$, $N_{bs}=64$, $P=0.5$, $\alpha=0.01$, $\lambda=0.05$, $\beta_1=0.9$, $\beta_2=0.999$, $\epsilon=10^{-8}$, and $l(k)$ and $u(k)$ are defined in (\ref{eq:lk}) and (\ref{eq:uk}), respectively. Clearly, \texttt{MBGD-RDA} is stable with respect to each of the four hyper-parameters in a wide range, which is desirable.

\subsection{Comparison with ANFIS} \label{sect:ANFIS}

ANFIS \cite{Jang1993} is an efficient algorithm for training TSK fuzzy systems on small datasets. This subsection compares the performance of \texttt{MBGD-RDA} with ANFIS on the first six smaller datasets.

\begin{figure}[htbp]         \centering
\includegraphics[width=\linewidth,clip]{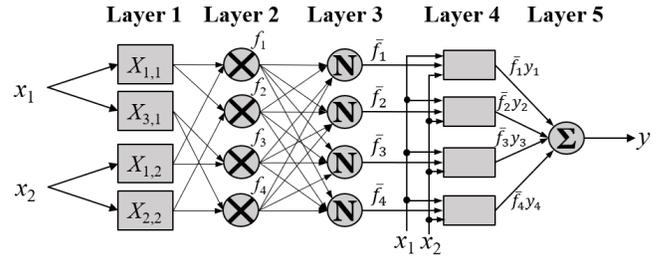}
\caption{A TSK fuzzy system represented as a 5-layer ANFIS. Note that $X_{1,1}=X_{2,1}$, $X_{3,1}=X_{4,1}$, $X_{1,2}=X_{3,2}$ and $X_{2,2}=X_{4,2}$ are used.} \label{fig:ANFIS}
\end{figure}

The ANFIS structure of a two-input one-output TSK fuzzy system is shown in Fig.~\ref{fig:ANFIS}. It has five layers:
\begin{itemize}
\item \emph{Layer 1:} The membership grade of $x_m$ on $X_{r,m}$ is computed, by (\ref{eq:mu}).
\item \emph{Layer 2:} The firing level of each rule $\mathrm{Rule}_r$ is computed, by (\ref{eq:fr}).
\item \emph{Layer 3:} The normalized firing levels of the rules are computed, by (\ref{eq:f}).
\item \emph{Layer 4:} Each normalized firing level is multiplied by its corresponding rule consequent.
\item \emph{Layer 5:} The output is computed, by (\ref{eq:yTSK2}).
\end{itemize}
All parameters of the ANFIS, i.e., the shapes of the MFs and the rule consequents, can be trained by GD \cite{Jang1993}. Or, to speed up the training, the antecedent parameters can be tuned by GD, and the consequent parameters by LSE \cite{Jang1993}.

In the experiment, we used the \emph{anfis} function in Matlab 2018b, which does not allow us to specify a batch size, but to use all available training examples in each iteration. For fair comparison, in \texttt{MBGD-RDA} we also set the batch size to the number of training examples. \emph{anfis} in Matlab has two optimization options: 1) batch GD for both antecedent and consequent parameters (denoted as \texttt{ANFIS-GD} in the sequel); and, 2) batch GD for antecedent parameters, and LSE for consequent parameters (denoted as \texttt{ANFIS-GD-LSE} in the sequel). We compared \texttt{MBGD-RDA} with both options.

\begin{figure}[htpb] \centering
\subfigure[]{\label{fig:ANFIS2}     \includegraphics[width=.98\linewidth,clip]{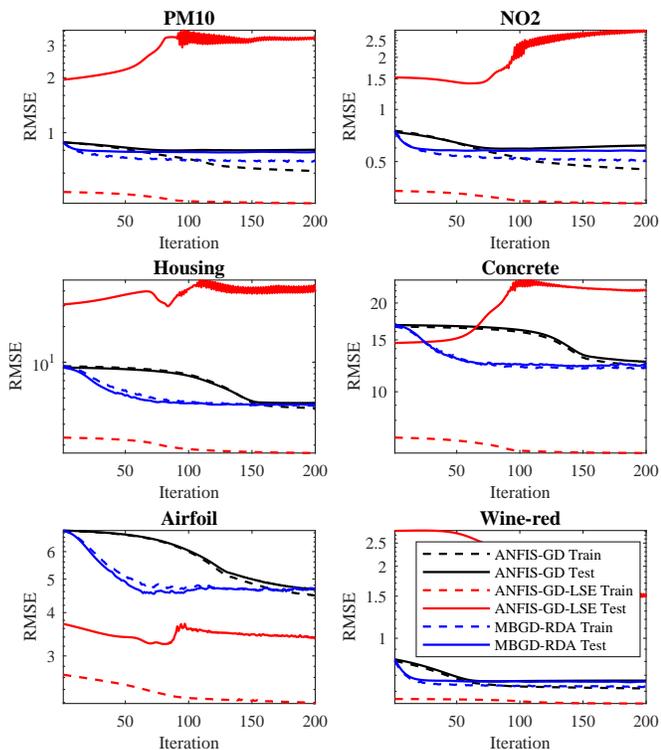}}
\subfigure[]{\label{fig:learningRate}     \includegraphics[width=.98\linewidth,clip]{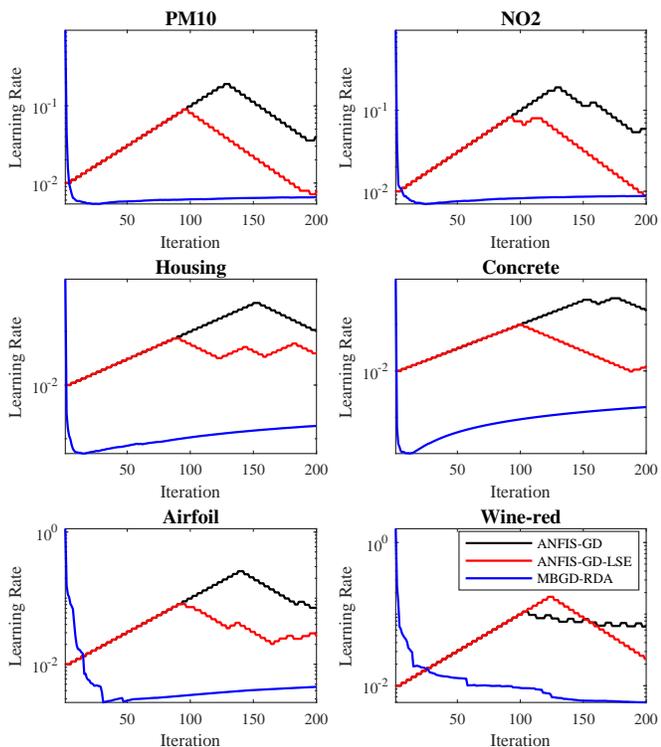}}
\caption{(a) Performance comparison of \texttt{MBGD-RDA} with \texttt{ANFIS-GD} and \texttt{ANFIS-GD-LSE} in batch GD. We use logarithmic scale on the vertical axis to make the curves more distinguishable. (b) Learning rates of \texttt{ANFIS-GD} and \texttt{ANFIS-GD-LSE}, and the average learning rate (across all model parameters) of \texttt{MBGD-RDA}, in batch GD.}
\end{figure}

The training and test RMSEs, averaged over 10 runs, are shown in Fig.~\ref{fig:ANFIS2}. \texttt{MBGD-RDA} always converged much faster than \texttt{ANFIS-GD}, and its best test RMSE was also always smaller. Additionally, it should be emphasized that \texttt{MBGD-RDA} can be used also for big data, whereas \texttt{ANFIS-GD} cannot.

Interestingly, although \texttt{ANFIS-GD-LSE} always achieved the smallest training RMSE, its test RMSE was almost always the largest, and had large oscillations. This suggests that \texttt{ANFIS-GD-LSE} had significant overfitting. If we could reduce this overfitting, e.g., through regularization, then \texttt{ANFIS-GD-LSE} could be a very effective TSK fuzzy system training algorithm for small datasets. This is one of our future research directions.

Fig.~\ref{fig:learningRate} shows the learning rates of \texttt{ANFIS-GD}, \texttt{ANFIS-GD-LSE} and \texttt{MBGD-RDA}. For the first two ANFIS based approaches, all model parameters shared the same learning rate. However, in \texttt{MBGD-RDA}, different model parameters had different learning rates, and we show the average learning rate across all model parameters on each dataset. The learning rates in \texttt{ANFIS-GD} and \texttt{ANFIS-GD-LSE} first gradually increased and then decreased. Interestingly, the learning rate of \texttt{ANFIS-GD-LSE} was almost always smaller than that of \texttt{ANFIS-GD} when the number of iterations was large. The learning rate of \texttt{MBGD-RDA} was always very high at the beginning, and then rapidly decreased. The initial high learning rate helped \texttt{MBGD-RDA} achieve rapid convergence.

\subsection{Comparison with DropMF and DropMembership}

In addition to DropRule, there could be other DropOut approaches in training a TSK fuzzy system, e.g.,
\begin{enumerate}
\item \emph{DropMF}, in which each input MF is dropped with a probability $1-P$, as illustrated in Fig.~\ref{fig:dropMF}. Dropping an MF is equivalent to setting the firing level of that MF to 1 (instead of 0). Comparing DropMF in Fig.~\ref{fig:dropMF2} and DropRule in Fig.~\ref{fig:dropRule2}, we can observe that each DropRule operation reduces the number of used rules by one; on the contrary, DropMF does not reduce the number of used rules; instead, it reduces the number of antecedents in multiple rules by one.

\begin{figure}[htpb] \centering
\subfigure[]{\label{fig:dropMF1}     \includegraphics[width=.9\linewidth,clip]{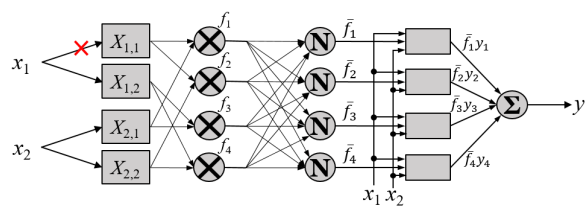}}
\subfigure[]{\label{fig:dropMF2}     \includegraphics[width=.9\linewidth,clip]{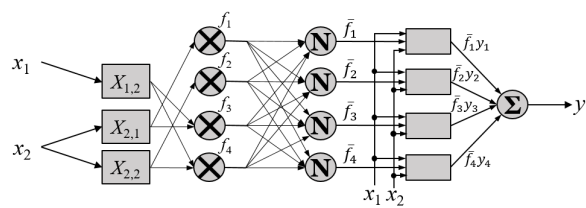}}
\caption{DropMF, where $X_{m,i}$ is the $i$th MF in the $m$th input domain. (a) The red cross indicates that MF $X_{1,1}$ for $x_1$ will be dropped; (b) the equivalent fuzzy system after dropping  $X_{1,1}$.}\label{fig:dropMF}
\end{figure}

\item \emph{DropMembership}, in which the membership of an input in each MF is dropped with a probability $1-P$, as illustrated in Fig.~\ref{fig:dropM}. Dropping a membership is equivalent to setting that membership to 1 (instead of 0). Comparing DropMembership in Fig.~\ref{fig:dropM2} and DropMF in Fig.~\ref{fig:dropMF2}, we can observe that DropMembership has a smaller impact on the firing levels of the rules than DropMF. For example, in Fig.~\ref{fig:dropMF2}, both $f_1$ and $f_2$ are impacted by DropMF, whereas in Fig.~\ref{fig:dropM2}, only $f_1$ is impacted by DropMembership.

\begin{figure}[htpb] \centering
\subfigure[]{\label{fig:dropM1}     \includegraphics[width=.9\linewidth,clip]{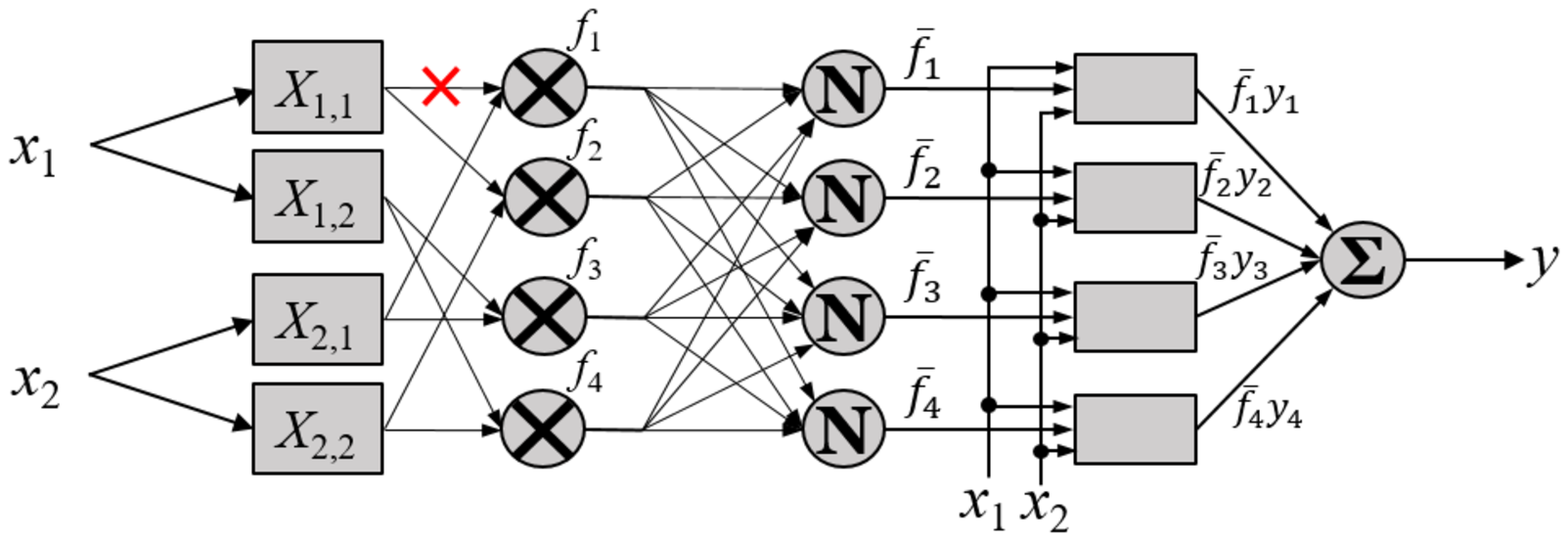}}
\subfigure[]{\label{fig:dropM2}     \includegraphics[width=.9\linewidth,clip]{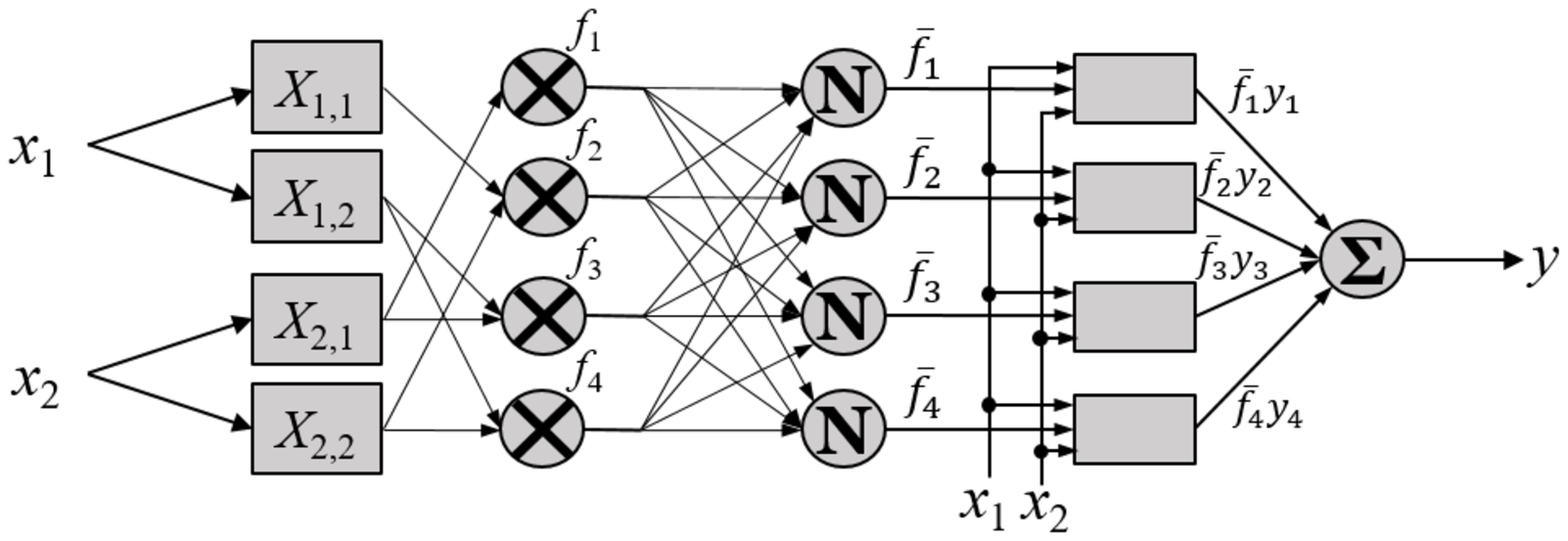}}
\caption{DropMembership, where $X_{m,i}$ is the $i$th MF in the $m$th input domain. (a) The red cross indicates that membership $\mu_{X_{1,1}}(x_1)$ in the first rule will be dropped; (b) the equivalent fuzzy system after dropping $\mu_{X_{1,1}}(x_1)$ in the first rule.}\label{fig:dropM}
\end{figure}

\end{enumerate}

\begin{figure*}[htpb]\centering
\includegraphics[width=.9\linewidth,clip]{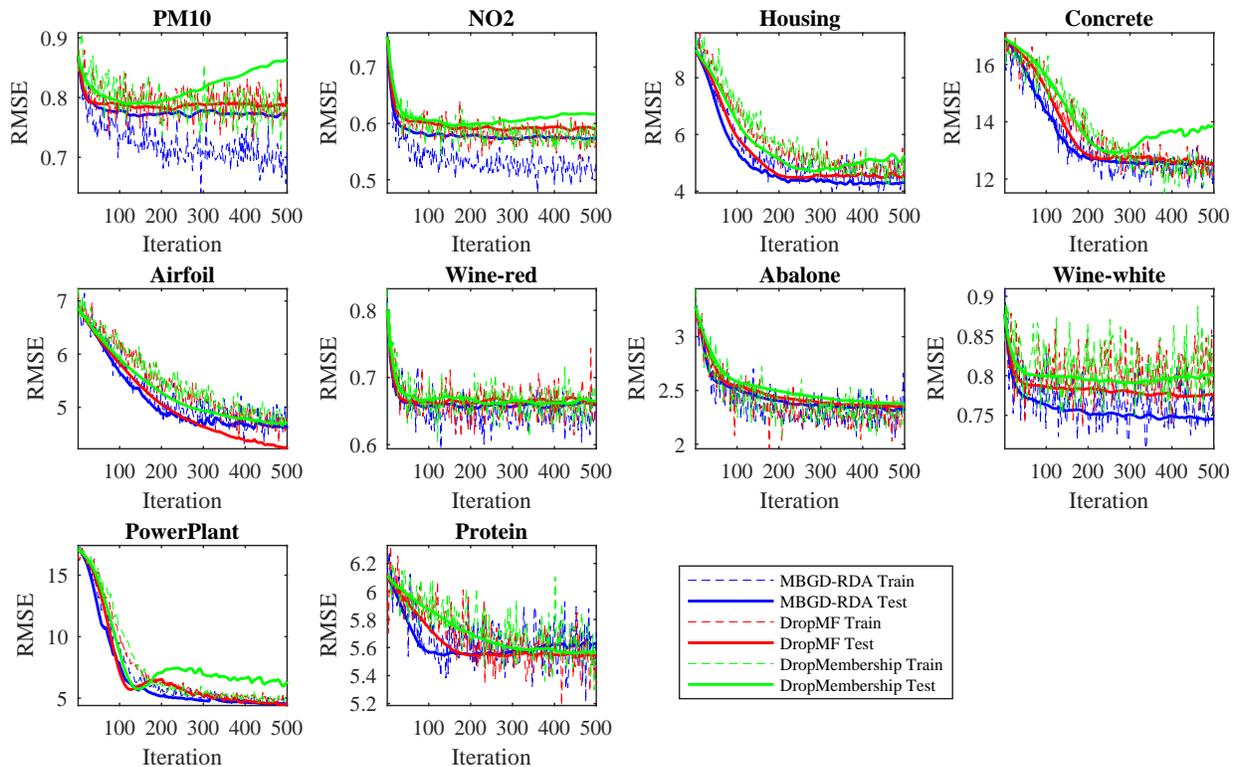}
\caption{The average training and test RMSEs of \texttt{MBGD-RDA}, \texttt{DropMF} (replacing DropRule in \texttt{MBGD-RDA} by DropMF) and \texttt{DropMembership} (replacing DropRule in \texttt{MBGD-RDA} by DropMembership) on the 10 datasets.} \label{fig:Drop}
\end{figure*}

Next, we compare the performances of DropMF, DropMembership with DropRule, by replacing DropRule in \texttt{MBGD-RDA} by DropMF and DropMembership, respectively. The training and test RMSEs, averaged over 10 runs, are shown in Fig.~\ref{fig:Drop}. Generally, DropRule performed the best, and DropMembership the worst. Comparing Figs.~\ref{fig:dropRule2}, \ref{fig:dropMF2} and \ref{fig:dropM2}, we can observe that DropRule introduces the maximum change to the TSK fuzzy system structure, and DropMembership the smallest. This suggests that a dropout operation that introduces more changes to the TSK fuzzy system may be more beneficial to the training and test performances.

\subsection{Comparison with Adam}

We also compared the performances of AdaBound with Adam. The learning algorithm, \texttt{MBGD-RD-Adam}, was identical to \texttt{MBGD-RDA}, except that AdaBound was replaced by Adam, by setting $l(k)=0$ and $u(k)=+\infty$ in Algorithm~\ref{alg:MBGD-RDA}.

\begin{figure*}[htpb]\centering
\includegraphics[width=.9\linewidth,clip]{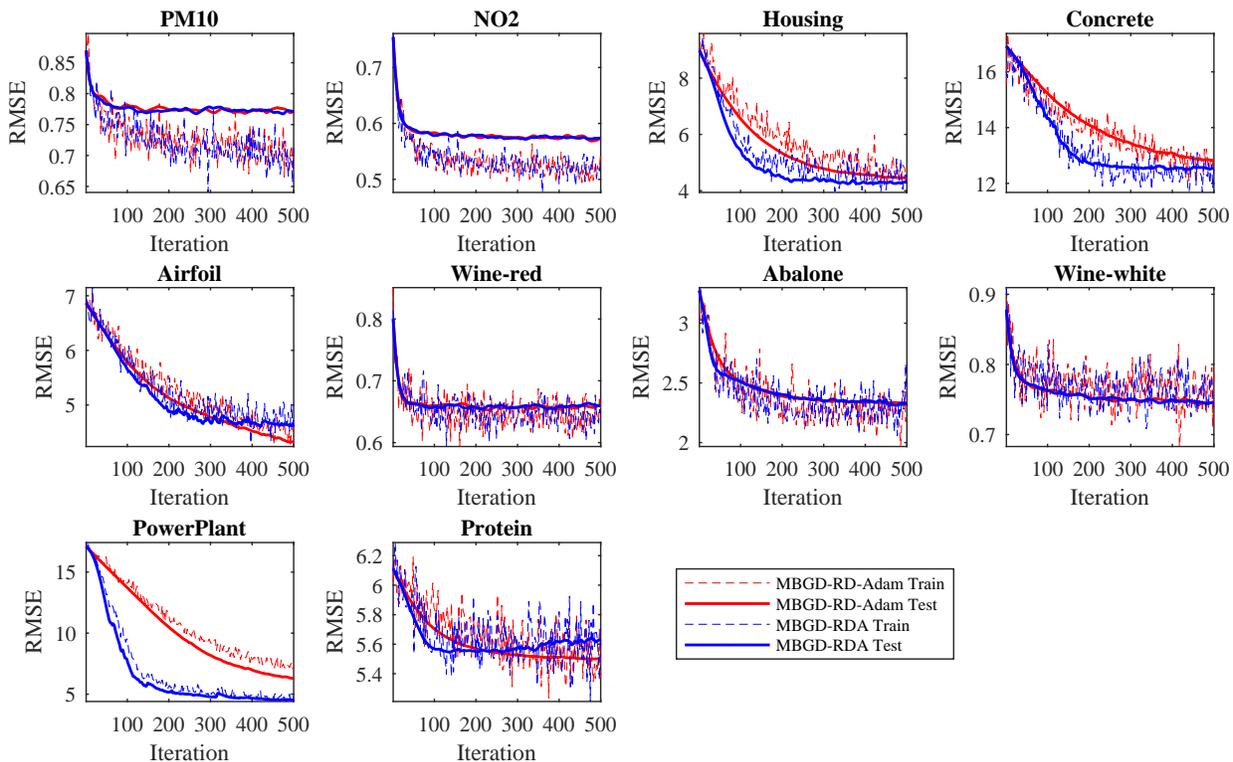}
\caption{The average training and test RMSEs of \texttt{MBGD-RDA} and \texttt{MBGD-RD-Adam} on the 10 datasets.} \label{fig:Adam}
\end{figure*}

The training and test RMSEs, averaged over 10 runs, are shown in Fig.~\ref{fig:Adam}. \texttt{MBGD-RDA} converged faster than, or equally fast with, \texttt{MBGD-RD-Adam}, and had smaller or comparable best test RMSEs as \texttt{MBGD-RD-Adam} on most datasets. So, it is generally safe to choose AdaBound over Adam.

\section{Conclusion and Future Research} \label{sect:conclusion}

TSK fuzzy systems are very useful machine learning models for regression problems. However, to our knowledge, there has not existed an efficient and effective training algorithm that enables them to deal with big data. Inspired by the connections between TSK fuzzy systems and neural networks, this paper extended three powerful optimization techniques for neural networks, e.g., MBGD, regularization, and AdaBound, to TSK fuzzy systems, and also proposed three novel techniques (DropRule, DropMF, and DropMembership) specifically for training TSK fuzzy systems. Our final algorithm, MBGD-RDA, which integrates MBGD, regularization, AdaBound and DropRule, can achieve fast convergence in training TSK fuzzy systems, and also superior generalization performance in testing. It can be used for training TSK fuzzy systems on datasets of any size; however, it is particularly useful for big datasets, for which currently no other efficient training algorithms exist. We expect that our algorithm will help promote the applications of TSK fuzzy systems, particularly to big data.

Finally, we need to point out that we have not considered various uncertainties in the data, e.g., missing values, wrong values, noise, outliers, etc., which frequently happen in real-world applications. Some techniques, e.g., rough sets \cite{Pawlak2012}, could be integrated with fuzzy sets to deal with them. Or, the type-1 TSK fuzzy systems used in this paper could also be extended to interval or general type-2 fuzzy systems \cite{Mendel2017,drwuFundamental2012} to cope with more uncertainties. These are some of our future research directions.


\end{document}